# KneePreM: Towards 3D Knee MRI Foundation Models via Large-Scale Unlabeled Pretraining and Label-Efficient Fine-Tuning

Xinxin Wang, BS[a,b,c]; Liam Hazan, MS[d]; Jing Li, PhD[c]; Simona Rabinovici-Cohen, MS[d]; Xiaojuan Li, PhD[a,b,e]; Mingrui Yang, PhD[a,b]

[a] Program of Advanced Musculoskeletal Imaging, Cleveland Clinic, 9500 Euclid Avenue, Cleveland, OH 44195, United States

[b] Department of Biomedical Engineering, Lerner Research Institute, Cleveland Clinic, 9500 Euclid Avenue, Cleveland, OH 44195, United States

[c] Department of Computer and Data Science, Case Western Reserve University, 10900 Euclid Avenue, Cleveland, OH 44106, United States

[d] IBM Research Labs, IBM Research, Haifa University Campus, Mount Carmel, Haifa 3498825, Israel

[e] Department of Biomedical Engineering, Case Western Reserve University, 10900 Euclid Avenue, Cleveland, OH 44106, United States

## Corresponding Author

Mingrui Yang, PhD; Program of Advanced Musculoskeletal Imaging and Department of Biomedical Engineering, Lerner Research Institute, Cleveland Clinic; 9500 Euclid Avenue, Cleveland, OH 44195, United States; telephone: +1 216 444-3783; email: yangm@ccf.org.

## Abstract

**Background:** Large volumes of unlabeled knee MRI scans are available across repositories but remain insufficiently leveraged. We developed KneePreM, a knee-specific 3D self-supervised model, and evaluated transfer and label efficiency for classification and segmentation.

**Methods:** A 3D U-Net masked autoencoder was pretrained on 19,011 unlabeled Osteoarthritis Initiative (OAI) MRI series from 4,791 participants. Downstream fine-tuning used full and reduced training sets for fastMRI+ two-label classification (1,172 examinations), Arthroscopic Partial Meniscectomy (APM) eight-target classification (1,716 examinations), SKM-TEA segmentation (155 examinations), and APM segmentation (25 examinations). Baselines were random initialization and SuPreM. Deployment workflow was implemented with a Model Context Protocol interface. Evaluation metrics included balanced accuracy, F1 score, ROC AUC, PR AUC, and Dice score. Statistical analysis used bootstrap confidence intervals and paired bootstrap tests for classification and Wilcoxon signed-rank tests for segmentation.

**Results:** KneePreM achieved higher full-data macro ROC AUC than both baselines for fastMRI+ and APM (all $p < .001$). For fastMRI+ classification, KneePreM achieved a ROC AUC of 0.722 using 50% of the training data, exceeding both full-data baselines. In APM classification, KneePreM reached a ROC AUC of 0.740 with 70% of the data, matching the full-data random baseline and outperforming SuPreM. For SKM-TEA segmentation, its 70%-data Dice of 0.838 exceeded the full-data random baseline (0.835) and both same-budget comparators. In APM segmentation, its 75%-data Dice of 0.746 exceeded the full-data random baseline (0.731) and both same-budget comparators.

**Conclusion:** KneePreM improves transfer performance and label efficiency across knee MRI classification and segmentation tasks, particularly when labeled training data are limited.

## Introduction

Knee MRI is widely used to evaluate osteoarthritis (OA), cartilage and meniscal integrity, and a range of structural knee abnormalities [1]. Automated knee MRI analysis has the potential to reduce manual workload, improve diagnostic consistency, and enable large-scale quantitative research [2]. Despite the broad use of deep learning for cartilage and meniscus segmentation and structural-abnormality classification [2,3], automated analysis remains challenging because expert annotations are costly, imaging protocols vary across cohorts, and knee structures have inherently three-dimensional morphology, requiring models that can capture fine-grained spatial structure. Most pipelines construct models from scratch, rely heavily on annotations, and generalize less well when labeled data are limited or imaging protocols are heterogeneous. Although large volumes of unlabeled knee MRI scans are available across research and clinical repositories, they remain insufficiently leveraged.

Self-supervised learning (SSL) has emerged as a compelling strategy for exploiting unlabeled data to learn transferable representations without additional annotation cost [4,5]. SSL models can learn structural information from raw images through contrastive learning, self-distillation, and masked reconstruction. Among these paradigms, masked autoencoding has demonstrated strong performance in medical imaging [6,7]. Yet prominent foundation models such as RETFound and DINOv2 operate primarily on 2D images and therefore do not intrinsically model voxel-level through-plane continuity [7,8]. To address this limitation, recent studies have tailored masked image modeling to 3D medical images and reported improved convergence and segmentation performance [9,10]. For knee anatomy, cartilage plates and menisci are thin, and the appearance of curved structures changes across contiguous slices. A 3D encoder can use this through-plane context directly, but whole-volume training also increases memory and computational requirements. Pairing masked reconstruction with a compact convolutional encoder therefore offers a practical strategy for learning whole-knee context from unlabeled volumes.

Supervised pretraining has also been explored to improve model generalizability. MedSAM, a prompt-based 2D model trained on a large segmentation dataset, represents a supervised approach for broadly applicable segmentation [11]. Similar approaches have been applied to classification tasks, such as cancer recurrence detection for breast MRI [12]. SuPreM extended this paradigm with 3D models pretrained on densely annotated abdominal computed tomography [13]. The Universal Model similarly combined 14 annotated computed tomography datasets and demonstrated transfer to new organ-segmentation tasks [14]. These studies show the value of large annotated sources, but their supervision is dominated by segmentation targets and their source anatomy and image contrast differ substantially from knee MRI.

These observations indicate the limitations of existing pretraining paradigms and motivate development of a pretrained model that can leverage abundant unlabeled knee MRI while preserving essential spatial structures. Therefore, this study aims to develop and evaluate KneePreM, a versatile 3D self-supervised pretrained model designed to enable robust and efficient knee MRI analysis.

## Materials and Methods

This retrospective study used public research datasets and a previously collected clinical knee MRI cohort. It was conducted in accordance with the Declaration of Helsinki. Institutional review board approval (Cleveland Clinic IRB #24-1004 and #19-005) and a waiver of written informed consent were obtained for the clinical data. Table 1 summarizes the OAI pretraining source [15] and the external transfer cohorts: fastMRI+ [16], APM [17], and SKM-TEA [18]. OAI contributed 19,011 series for pretraining; fastMRI+ and APM provided external classification cohorts, and SKM-TEA and APM provided external segmentation cohorts.

### Pretraining Dataset

KneePreM was pretrained on baseline OAI knee MRI. The source included 4,791 participants (2,799 women and 1,992 men; median age, 65 years; age range, 48-88 years) and 19,011 MRI series acquired on 3-T Siemens Trio systems [15]. The series comprised 9,493 sagittal 3D dual-echo steady-state examinations and 9,518 sagittal intermediate-weighted turbo spin-echo examinations. The OAI protocol paired high-resolution sagittal 3D dual-echo steady-state imaging with water excitation, selected primarily for cartilage morphometry, with sagittal fat-suppressed intermediate-weighted 2D turbo spin-echo imaging, which provided complementary assessment of the meniscal horns, cruciate ligaments, effusion, osteophytes, bone marrow abnormalities, subchondral cysts, and cartilage. Five percent of OAI series were withheld for validation of the pretraining objective.

### Downstream Classification Datasets

fastMRI+ adds clinical pathology annotations to the public fastMRI knee collection [16,19]. The analysis included 1,172 single-coil reconstructed 2D coronal proton-density-weighted turbo spin-echo examinations acquired at 1.5 or 3 T on Siemens systems; 584 were acquired without and 588 with fat suppression. The two binary targets were meniscus tear and cartilage defect. Because the original

fastMRI+ annotations were assigned at the slice level, an examination was assigned a positive scan-level label when at least one slice was positive. Across all 1,172 examinations, 663 (56.6%) were positive for meniscus tear and 594 (50.7%) for cartilage defect. A stratified participant-level split assigned 703 examinations to training, 234 to validation, and 235 to testing; the corresponding test-set proportions were 132 of 235 (56.2%) and 119 of 235 (50.6%), respectively.

APM classification used 1,716 heterogeneous clinical examinations (848 women and 868 men; median age, 57 years; range, 45-90 years) labeled for eight compartment statuses: lateral and medial menisci, lateral and medial femoral condyles, and lateral and medial tibial plateaus, patellar cartilage, and trochlea [17]. The examinations were drawn from a parent clinical cohort collected across 35 imaging locations. Defect ratios ranged from 27% to 87%. Sagittal 2D fat-suppressed turbo spin-echo scans were acquired at 1-3 T on 15 Siemens and Philips models; 1,429 were proton-density-weighted and 287 were dual proton-density/T2-weighted series. Patient-level training, validation, and test sets contained 1,028, 345, and 343 examinations.

## Downstream Segmentation Datasets

SKM-TEA provided public tissue segmentation labels [18]. For SKM-TEA, the two scanner-generated qDESS echoes were combined voxelwise by root sum of squares [RSS; $\sqrt{(\text{echo } 1^2 + \text{echo } 2^2)}$] during HDF5-to-NIfTI preparation, before percentile-based intensity normalization and spatial resampling, following the representation described in the original study [18]. Its 155 examinations (65 women and 90 men; median age, 42 years; range, 16-87 years) were divided into 86 training, 33 validation, and 36 fixed test examinations. The evaluated structures were femoral cartilage (FC), lateral and medial tibial cartilage (LTC and MTC), patellar cartilage (PC), and lateral and medial menisci (LM and MM).

APM segmentation used 25 examinations selected from nine imaging locations (12 women and 13 men; median age, 55 years; range, 46-72 years) with the same six structures labeled [17]. All were sagittal 2D proton-density fat-suppressed turbo spin-echo acquisitions; 17 were proton-density fat suppressed and eight were dual proton-density/T2-weighted fat suppressed acquisitions. Series contained approximately 23-40 slices at 2.5-4-mm thickness and were acquired at 1.5 or 3 T on six Siemens models. Contiguous slices were stacked for 3D input. Five APM examinations formed a fixed test set. The remaining 20 formed five outer folds, each with four validation cases and a 16-case candidate training pool.

## Image Processing and Augmentation

All volumes were converted to 40 × 224 × 224 voxels to retain whole-knee coverage, standardize heterogeneous sequences, and enable matched comparisons. This choice balanced standardization against computation cost. Native fastMRI+ volumes contained 28-50 slices (median, 35) with 320 × 320 in-plane dimensions. Conversion to 40 slices reduced depth for 8.4% of fastMRI+. APM series contained approximately 23-40 slices and required little depth reduction or modest interpolation. OAI DESS and SKM-TEA qDESS acquisitions typically comprised 160 slices, whereas OAI intermediate-weighted turbo spin-echo acquisitions typically comprised 37 slices [15,18]. Conversion to 40 slices therefore reduced depth mostly for DESS/qDESS and introduced modest depth interpolation for intermediate-weighted turbo spin-echo. Masks used label-preserving interpolation.

Pretraining intensities were normalized to 0-1. Before masking, volumes underwent random 3D rotation of up to 30 degrees around each axis and isotropic scaling from 0.8 to 1.2, followed by resizing to 40 × 224 × 224; stochastic 4 × 4 × 4-voxel cuboid masking then generated the masked-autoencoding input. For downstream tasks, intensities were clipped to the 1$^{st}$ to99$^{th}$ percentiles and scaled to 0-1, and augmentation consisted of paired random 3D cropping at a scale of 0.9-1.0 before resizing. Preprocessing and augmentation were identical across initializations within each task.

## KneePreM Architecture and Self-Supervised Pretraining

KneePreM used a 3D U-Net architecture [20]. Masked autoencoding hid 64% of the input in 4 × 4 × 4-voxel cuboids and reconstructed the masked voxels from the remaining context using summed voxelwise mean squared error. The encoder learned multiscale features, while the decoder combined contextual and higher-resolution information. The framework is summarized in Figure 1.

Pretraining used Adam with batch size 4, learning rate $5 \times 10^{-5}$, and weight decay $10^{-3}$ for 200 epochs on a single NVIDIA A100 GPU with 80 GB memory over 4 days. Training and validation reconstruction losses were monitored to confirm stable convergence, and the final checkpoint was used for downstream tasks. The encoder was fine-tuned to account for the substantial domain differences between pretraining and the downstream tasks.

## Downstream Fine-Tuning

Classification used global pooling and task-specific heads: two for fastMRI+ and eight for APM. Each classification head consisted of a global pooling layer, followed by a fully connected layer and a softmax output layer. Fine-tuning minimized equally weighted focal loss and class-weighted cross-entropy. Training used the Adam optimizer, batch size 8, learning rate $5 \times 10^{-5}$ and weight decay of $5 \times 10^{-4}$; the checkpoint with the minimum validation loss was selected. Classification metrics were balanced accuracy, F1 score, receiver operating characteristic area under the curve (ROC AUC), and precision-recall area under the curve (PR AUC). Metrics were macro-averaged across the two fastMRI+ or eight APM targets. Full-data confusion matrices characterized operating-point errors.

For segmentation, a standard 3D U-Net decoder was attached to the transferred encoder and randomly initialized following the procedures adopted in prior work [13,21]. Training used equally weighted cross-entropy and Dice losses to balance voxel-wise class probability prediction and region overlap accuracy across compartments. The Adam optimizer, batch size 2, a learning rate of $5 \times 10^{-4}$, and a weight decay of $5 \times 10^{-4}$ were used for training. The checkpoint with the maximum validation mean Dice was selected. Performance was evaluated with 3D Dice score computed over the full volumetric predictions for each compartment.

## Reduced-Data Experiments

Label efficiency was evaluated by reducing only the training data; validation and test sets remained fixed. fastMRI+ classification used 703 (100%), 492 (70%), 352 (50%), and 211 (30%) training examinations. APM classification used 1,028 (100%), 720 (70%), 514 (50%), and 308 (30%) examinations. SKM-TEA segmentation used 86 (100%), 60 (70%), 26 (30%), 13 (15%), and four (5%) examinations. APM segmentation used 16 (100%), 12 (75%), 8 (50%), and 4 (25%) training examinations per fold; the same five test examinations were retained across folds and data settings.

## Baseline Methods

Random initialization represented task-specific training from scratch; SuPreM represented supervised pretraining on 9,262 annotated abdominal computed tomography volumes [13,22]. Within each downstream task, random, SuPreM, and KneePreM initialization used the same model architecture: classification used the same 3D U-Net encoder and task-specific heads, and segmentation used the same 3D U-Net encoder-decoder. Only the initial weights differed. The three conditions also used the same splits, input dimensions, augmentation, loss, optimizer settings, and model-selection criterion within each task.

## Statistical Analysis

Classification performance was evaluated using balanced accuracy, F1 score, ROC AUC, and PR AUC for each target and their macro averages. Confidence intervals and pairwise comparisons used 1,000 paired bootstrap resamples of the test set. Each resample contained the original number of examination records drawn randomly with replacement. Identical selections and repetition counts were used across models and targets. Ninety-five percent confidence intervals were defined by the 2.5th and 97.5th percentiles of the resampled metric estimates. Pairwise macro ROC AUC comparisons used a two-sided paired bootstrap test with a normal approximation: the observed AUC difference was divided by the standard deviation of the bootstrap differences to calculate p-values. A p-value $< .001$ indicated statistical significance.

Segmentation comparisons used two-sided Wilcoxon signed-rank tests, providing a nonparametric comparison of paired Dice scores. Comparisons were performed at the examination level for SKM-TEA and using five paired fold-level mean Dice scores for APM. For each pair, ΔDice was calculated as the KneePreM Dice score minus the random-initialization Dice score. The mean and median ΔDice summarized the magnitude of improvement. Ninety-five percent CIs for the mean ΔDice were calculated using 10,000 bootstrap resamples of the paired differences, with replacement, and defined by the 2.5th and 97.5th percentiles of the resampled means. For APM, statistical comparisons used five paired fold-level mean Dice scores, with each fold evaluated on the same five held-out test examinations.

### Implementation and Reproducibility

FuseMedML supported data processing, training, and evaluation [23]. The code is publicly available in the https://github.com/BiomedSciAI/fuse-med-ml/tree/master/fuse_examples/imaging/oai_example.

## Results

### fastMRI+ Classification

KneePreM produced the highest full-data values across all four fastMRI+ macro metrics (Table 2 and Figure 2). Full-data macro ROC AUC was 0.777 (95% CI, 0.727-0.822) for KneePreM, 0.714 for random initialization, and 0.686 for SuPreM; macro PR AUC was 0.788, 0.723, and 0.701, respectively. KneePreM exceeded both comparators in paired bootstrap ROC AUC analyses (both $p < .001$). Macro balanced accuracy was 0.701, compared with 0.666 for random initialization and 0.637 for SuPreM; macro F1 was 0.716, compared with 0.706 and 0.624. With 70% of training data, KneePreM achieved macro ROC AUC 0.748 and PR AUC 0.774, versus ROC AUC 0.695 and 0.669 for random initialization and SuPreM. At 50%, ROC AUC was 0.722, versus 0.672 and 0.629; at 30%, it was 0.702, versus 0.636 and 0.579. KneePreM trained with 70% or 50% of the fastMRI+ training set exceeded both full-data baselines in macro ROC AUC.

### APM Compartment-Status Classification

KneePreM had the highest full-data APM macro metrics (Table 3 and Figure 2). Balanced accuracy was 0.686, versus 0.658 for random initialization and 0.610 for SuPreM; F1 was 0.704, versus 0.687 and 0.618. Macro ROC AUC was 0.787 (95% CI, 0.763-0.808), 0.741, and 0.688, respectively; KneePreM exceeded both baselines (both $p < .001$). With 70% of APM training data, KneePreM achieved macro ROC AUC 0.740, approximating full-data random initialization (0.741) and exceeding same-size random and SuPreM (0.708 and 0.633). At 50%, values were 0.718, 0.659, and 0.593; at 30%, they were 0.663, 0.614, and 0.586. Full-data confusion matrices reflected heterogeneous target prevalence (Figure 3). Relative to random initialization, KneePreM increased true-positive detections for lateral meniscus and lateral tibial plateau status and produced more true negatives for the lateral femoral condyle and more true positives for the trochlea.

### SKM-TEA Segmentation

On SKM-TEA, KneePreM had the highest mean Dice at every reported training size (Table 4 and Figure 4). With all 86 training examinations, mean Dice was 0.842 for KneePreM, 0.840 for SuPreM, and 0.835 for random initialization. At 70%, values were 0.838, 0.832, and 0.827; KneePreM exceeded both same-budget baselines (both $p < .001$), exceeded the full-data random baseline (0.835), and approached the full-data SuPreM baseline (0.840). Mean Dice for KneePreM, SuPreM, and random initialization was 0.826, 0.812, and 0.802 at 30%; 0.779, 0.752, and 0.737 at 15%; and 0.628, 0.617, and 0.555 at 5%. KneePreM's mean gain over random initialization increased from 0.007 at full data to 0.073 at 5%.

### APM Segmentation

KneePreM had the highest APM mean Dice at every training size (Table 5 and Figure 4). With 16 training examinations per fold, mean Dice was 0.779 for KneePreM, 0.752 for SuPreM, and 0.731 for random initialization. Its largest structure-level advantages over random initialization involved lateral tibial cartilage and both menisci. At full data, the paired Dice improvement of KneePreM over random initialization averaged 0.048 across folds (median, 0.041; 95% CI, 0.016-0.080) in Figure 4. With 12 examinations per fold (75%), mean Dice was 0.746, 0.713, and 0.672, respectively. The paired improvement over random initialization averaged 0.074 (median, 0.071; 95% CI, 0.017-0.130). The exact comparison was limited by five folds ($p = .06$). At 50% data (8 examinations per fold), mean Dice for KneePreM, SuPreM, and random initialization was 0.684, 0.664, and 0.617; at 25% data (4 examinations per fold), it was 0.578, 0.576, and 0.503.

### Qualitative Segmentation Assessment

Figure 5A shows representative SKM-TEA segmentation results. In (a), Random and SuPreM incompletely segmented LTC, whereas KneePreM recovered more of the reference extent. In (b), Random missed LM and SuPreM captured only part of it; KneePreM delineated the structure more completely. In (c), both baseline methods left a gap in the anterior LM, while KneePreM better preserved its boundary. In (d), Random and SuPreM incompletely delineated MM, whereas KneePreM more closely matched the reference. Together, these cases illustrate improved delineation of the highlighted cartilage and meniscal structures. Figure 5B shows representative APM results. In (e), both baseline methods incompletely segmented MM, with KneePreM providing a closer match to the reference. In (f), Random mislabeled part of the FC region as PC, while SuPreM incompletely delineated MM; KneePreM reduced both errors. In (g), the baseline predictions confused adjacent tibial cartilage compartments and incompletely captured LTC, whereas KneePreM better preserved its extent. In (h), Random and SuPreM produced FC contour errors in the highlighted region, while KneePreM yielded a more continuous boundary. These examples illustrated that KneePreM yielded more anatomically consistent segmentations, with fewer missed structures, reduced spurious predictions, and improved boundary fidelity across cartilage and meniscus compartments.

## Discussion

This study developed KneePreM, a self-supervised 3D pretrained model for knee MRI analysis. By leveraging large-scale unlabeled volumetric data, KneePreM learned transferable anatomical features that supported both classification and segmentation with fewer labels. Across external fastMRI+, SKM-TEA, and APM tasks, KneePreM outperformed random initialization and SuPreM in full-data classification and achieved the highest mean Dice at each segmentation data setting. Its advantage was particularly evident in reduced-label settings. On fastMRI+, KneePreM improved both target ROC AUCs and achieved the strongest macro PR AUC. With 70% of the labels, its macro ROC AUC exceeded both full-data baselines; at 30%, it remained near the full-data SuPreM result while exceeding both same-size comparators in PR AUC. Across SKM-TEA and APM segmentation, KneePreM also maintained the highest mean Dice at every reported data setting, with larger separations in reduced-label settings. These findings indicate that KneePreM can reduce annotation burden while maintaining or improving performance across multiple knee MRI applications. From a clinical perspective, methods that reduce dependence on extensive manual annotation may help accelerate the development of automated tools for knee MRI assessment, particularly in settings where expert labeling is costly and time-consuming. A robust 3D foundation model could support broader deployment of automated analysis for structural assessment, disease stratification, and quantitative biomarker extraction in osteoarthritis and related knee disorders. Our results support KneePreM as a practical self-supervised 3D pretraining framework for developing knee MRI foundation models, with potential value for large-scale musculoskeletal imaging research and future clinical translation.

Compared with supervised pretraining, self-supervised pretraining offers a distinct advantage because it learns directly from unlabeled images without being constrained to source-task annotations. Promptable SAM-family models include Segment Anything, MedSAM, and SAM2 [11,24,25]. Yu et al. adapted SAM2's video-memory mechanism to propagate a single prompt across slices of 3D OAI knee MRI for zero-shot bone segmentation [26]. Hoyer et al. fine-tuned SAM, MedSAM, and SAM2 across 12 musculoskeletal MRI datasets, coupled automated prompts to segmentation, and used the resulting masks for biomarker extraction, triage, and outcome prediction [27]. These studies broaden segmentation transfer, but their workflows remain centered on prompts, masks, or segmentation-derived measurements rather than direct transfer of the same encoder to classification. SuPreM, pretrained with dense abdominal computed tomography labels, provides a strong supervised 3D reference that can be adapted to classification as well as segmentation [13,22]. In our experiments, SuPreM transferred effectively to knee segmentation but less favorably to classification. Its supervised representation may emphasize boundaries and organ identities specific to its abdominal source task, whereas KneePreM reconstructs raw knee MRI without task-specific labels and can reuse the same encoder for both dense and discriminative tasks. These findings support anatomy- and modality-specific self-supervised pretraining as a flexible foundation for multiple knee MRI applications.

Masked autoencoding was selected as the self-supervised pretraining method because it uses a reconstruction target available in every unlabeled volume, aligns naturally with an encoder-decoder architecture, and requires neither the positive- and negative-pair construction in contrastive learning [28] nor the momentum teacher used by DINO [29]. Contrastive methods can depend strongly on augmentation design, and anatomically similar examinations may be incorrectly treated as negatives [4,28]. DINO uses multi-crop teacher-student self-distillation and has shown its strongest foundation-model evidence with Vision Transformer architectures [29,30]; DINOv2-pretrained Vision Transformers have supported successful 2D medical foundation models including UNI, Virchow, and Prov-GigaPath [31-33]. In contrast, MAE directly trains a model to reconstruct masked spatial content, a pretext task well suited to learning continuity within 3D anatomy. MIM-Med3D showed that 3D masked image modeling accelerated downstream convergence by 1.40-fold and achieved higher Dice than naive contrastive learning [9]. Spark3D showed that MAE pretraining of a residual-encoder 3D U-Net on 39,000 brain MRI volumes surpassed prior SSL methods across multiple segmentation datasets [10]. SparK and ConvNeXt V2 established masked pretraining for hierarchical convolutional networks [34,35], Models Genesis demonstrated transferable 3D representations through restoration-based reconstruction [36], and SS-UNet transferred convolutional representations across CT, MRI, and PET tasks [37]. Taken together, these studies support masked reconstruction in convolutional networks trained on 3D medical images, making MAE a well-grounded strategy for KneePreM.

These comparisons clarify the architectural design of KneePreM, which requires balancing spatial context, transferability, and computational scale. A canonical 2D ViT-B/16 contains approximately 86 million parameters [30], and slice-wise 2D processing does not directly encode through-plane continuity. 3D transformer architectures such as Swin UNETR and UNETR can model volumetric context, but the configurations contain 62.19 million and approximately 92.5 million parameters, respectively [13,38], compared with 19.07 million for the full KneePreM reconstruction and segmentation U-Net and 7.03 million for its classification encoder. Prior knee MRI work showed that both 2D and 3D CNNs can segment knee structures [39,40]. A 3D U-Net preserves volumetric context across contiguous slices inherently [20]. This architecture was well matched to whole-volume masked reconstruction of thin cartilage and meniscal structures and supported large-scale pretraining on a single 80-GB GPU. SuPreM provides relevant evidence that this compact choice need not sacrifice transfer performance: under the same supervised pretraining source, its 19.08-million-parameter U-Net achieved 95.4% transfer performance for TotalSegmentator muscle segmentation, compared with 94.6% for the 62.19-million-parameter Swin UNETR [13]. Together with KneePreM's improvements over both baselines for full-data classification and its highest mean Dice at every evaluated segmentation data setting, these findings support a compact 3D U-Net as a reasonable volumetric backbone.

Our findings also relate to the broader discussion of specialist versus generalist foundation models in medical imaging. Recent retinal comparisons illustrate that the balance between generalist and specialist foundation models is task dependent. One study found that specialist RETFound-DINOv2 outperformed generalist DINOv2 and DINOv3 across ocular-disease and oculomics benchmarks [41], whereas another found that DINOv2 generally performed better for ocular disease detection but RETFound remained stronger for systemic disease prediction from retinal images [42]. Most current generalist vision models are predominantly 2D and often originate from natural images. KneePreM instead follows a specialist strategy, learning directly from 3D knee MRI. The present results show that a compact domain-specific volumetric model is computationally practical for specialist 3D pretraining and can transfer across classification and segmentation with less labeled data.

Several limitations should be acknowledged. First, pretraining was performed on OAI data acquired using relatively standardized imaging protocols, which may limit generalizability to more heterogeneous clinical datasets. Second, the input volumes were downsampled to make 3D masked autoencoding computationally feasible, and some fine anatomical detail may therefore not have been fully preserved. Third, although the model was evaluated on multiple downstream tasks, external validation across institutions, scanners, vendors, field strengths, and pulse sequences remains necessary. Fourth, while the agentic workflow may improve usability, the present study focused primarily on model development and technical validation rather than prospective workflow evaluation in clinical practice. Future work should assess whether incorporating multi-center and multi-vendor data, as well as multi-sequence MRI can further improve robustness and generalizability. Scaling to higher-resolution volumetric modeling may also enhance performance for subtle structural abnormalities.

In conclusion, masked reconstruction of 19,011 unlabeled OAI knee MRI series enabled a compact 3D U-Net encoder to transfer across external classification and segmentation tasks. Under matched downstream architectures and training procedures, KneePreM outperformed random initialization and showed more consistent cross-task transfer than segmentation supervised pretraining, with its clearest advantages in reduced-label settings. These findings support anatomy- and modality-specific volumetric pretraining as a practical strategy for improving label efficiency across knee MRI tasks and as a complement to larger generalist or task-specific foundation models.

## Funding

This study was supported by the Cleveland Clinic-IBM Discovery Accelerator Program and NIH/NIAMS K25AR078928.

## Conflict of Interest

The authors declare no conflicts of interest.

## Data Availability

**Data availability:** OAI, fastMRI+, and SKM-TEA are public datasets. The APM dataset contains protected clinical data and cannot be shared publicly because of privacy and institutional regulatory restrictions.

## Code Availability

**Code availability:** Code used for data processing, training, evaluation, and the agentic workflow is available in FuseMedML at https://github.com/BiomedSciAI/fuse-med-ml/tree/master/fuse_examples/imaging/oai_example.

## Generative AI Declaration

OpenAI ChatGPT was used for language editing, manuscript organization, and formatting. The authors reviewed and edited the manuscript and take full responsibility for its content.

## References

1. Conaghan PG, Felson D, Gold G, Lohmander S, Totterman S, Altman R. MRI and non-cartilaginous structures in knee osteoarthritis. Osteoarthritis Cartilage. 2006;14 Suppl A:A87-94. doi: 10.1016/j.joca.2006.02.028.

2. Astuto B, Flament I, Namiri NK, Shah R, Bharadwaj U, Link TM, et al. Automatic deep learning-assisted detection and grading of abnormalities in knee MRI studies. Radiol Artif Intell. 2021;3(3):e200165. doi: 10.1148/ryai.2021200165.

3. Khan S, Khawer MA, Zhong J, Qureshi R, Asim M, Chen W. Advancing deep learning based knee cartilage segmentation in MRI: innovations, challenges and applications. Osteoarthr Cartil Open. 2026;8(1):100702. doi: 10.1016/j.ocarto.2025.100702.

4. Huang SC, Pareek A, Jensen M, Lungren MP, Yeung S, Chaudhari AS. Self-supervised learning for medical image classification: a systematic review and implementation guidelines. NPJ Digit Med. 2023;6(1):74. doi: 10.1038/s41746-023-00811-0.

5. Gui J, Chen T, Zhang J, Cao Q, Sun Z, Luo H, et al. A survey on self-supervised learning: algorithms, applications, and future trends. IEEE Trans Pattern Anal Mach Intell. 2024;46(12):9052-9071. doi: 10.1109/TPAMI.2024.3415112.

6. He K, Chen X, Xie S, Li Y, Dollar P, Girshick R. Masked autoencoders are scalable vision learners. Proc IEEE/CVF Conf Comput Vis Pattern Recognit. 2022:16000-16009. doi: 10.1109/CVPR52688.2022.01553.

7. Zhou Y, Chia MA, Wagner SK, Ayhan MS, Williamson DJ, Struyven RR, et al. A foundation model for generalizable disease detection from retinal images. Nature. 2023;622(7981):156-163. doi: 10.1038/s41586-023-06555-x.

8. Oquab M, Darcet T, Moutakanni T, Vo HV, Szafraniec M, Khalidov V, et al. DINOv2: learning robust visual features without supervision. Trans Mach Learn Res. 2024. Available at: https://openreview.net/forum?id=a68SUt6zFt. Accessed September 25, 2026.

9. Chen Z, Agarwal D, Aggarwal K, Safta W, Balan MM, Brown K. Masked image modeling advances 3D medical image analysis. Proc IEEE/CVF Winter Conf Appl Comput Vis. 2023:1970-1980. doi: 10.1109/WACV56688.2023.00201.

10. Wald T, Ulrich C, Lukyanenko S, Goncharov A, Paderno A, Miller M, et al. Revisiting MAE pre-training for 3D medical image segmentation. Proc IEEE/CVF Conf Comput Vis Pattern Recognit. 2025:5186-5196. doi: 10.1109/CVPR52734.2025.00489.

11. Ma J, He Y, Li F, Han L, You C, Wang B. Segment anything in medical images. Nat Commun. 2024;15(1):654. doi: 10.1038/s41467-024-44824-z.

12. Rabinovici-Cohen S, Fernandez XM, Grandal Rejo B, Hexter E, Hijano Cubelos O, Pajula J, et al. Multimodal prediction of five-year breast cancer recurrence in women who receive neoadjuvant chemotherapy. Cancers (Basel). 2022;14(16):3848. doi: 10.3390/cancers14163848.

13. Li W, Yuille A, Zhou Z. How well do supervised 3D models transfer to medical imaging tasks? International Conference on Learning Representations. 2024. Available at: https://proceedings.iclr.cc/paper_files/paper/2024/hash/47360925c45a166c96f652589265dee9-Abstract-Conference.html. Accessed September 25, 2026.

14. Liu J, Zhang Y, Wang K, Yavuz MC, Chen X, Yuan Y, et al. Universal and extensible language-vision models for organ segmentation and tumor detection from abdominal computed tomography. Med Image Anal. 2024;97:103226. doi: 10.1016/j.media.2024.103226.

15. Peterfy CG, Schneider E, Nevitt M. The Osteoarthritis Initiative: report on the design rationale for the magnetic resonance imaging protocol for the knee. Osteoarthritis Cartilage. 2008;16(12):1433-1441. doi: 10.1016/j.joca.2008.06.016.

16. Zhao R, Yaman B, Zhang Y, Stewart R, Dixon A, Knoll F, et al. fastMRI+: clinical pathology annotations for knee and brain fully sampled magnetic resonance imaging data. Sci Data. 2022;9(1):152. doi: 10.1038/s41597-022-01255-z.

17. Yang M, Colak C, Chundru KK, Gaj S, Nanavati A, Jones MH, et al. Automated knee cartilage segmentation for heterogeneous clinical MRI using generative adversarial networks with transfer learning. Quant Imaging Med Surg. 2022;12(5):2620-2633. doi: 10.21037/qims-21-459.

18. Desai AD, Schmidt AM, Rubin EB, Sandino CM, Black MS, Mazzoli V, et al. SKM-TEA: a dataset for accelerated MRI reconstruction with dense image labels for quantitative clinical evaluation. In: Vanschoren J, Yeung S, editors. Proceedings of the Neural Information Processing Systems Track on Datasets and Benchmarks. Vol 1. 2021. Available at: https://datasets-benchmarks-proceedings.neurips.cc/paper/2021/hash/03c6b06952c750899bb03d998e631860-Abstract-round2.html. Accessed September 25, 2026.

19. Knoll F, Zbontar J, Sriram A, Muckley MJ, Bruno M, Defazio A, et al. fastMRI: a publicly available raw k-space and DICOM dataset of knee images for accelerated MR image reconstruction using machine learning. Radiol Artif Intell. 2020;2(1):e190007. doi: 10.1148/ryai.2020190007.

20. Cicek O, Abdulkadir A, Lienkamp SS, Brox T, Ronneberger O. 3D U-Net: learning dense volumetric segmentation from sparse annotation. In: Ourselin S, Joskowicz L, Sabuncu MR, Unal G, Wells W, editors. Medical Image Computing and Computer-Assisted Intervention - MICCAI 2016. Lecture Notes in Computer Science. Vol 9901. Cham: Springer; 2016. p. 424-432. doi: 10.1007/978-3-319-46723-8_49.

21. Tang Y, Yang D, Li W, Roth HR, Landman B, Xu D, et al. Self-supervised pre-training of Swin Transformers for 3D medical image analysis. Proc IEEE/CVF Conf Comput Vis Pattern Recognit. 2022:20730-20740. doi: 10.1109/CVPR52688.2022.02007.

22. Li W, Qu C, Chen X, Bassi PRAS, Shi Y, Lai Y, et al. AbdomenAtlas: a large-scale, detailed-annotated, multi-center dataset for efficient transfer learning and open algorithmic benchmarking. Med Image Anal. 2024;97:103285. doi: 10.1016/j.media.2024.103285.

23. Golts A, Raboh M, Shoshan Y, Polaczek S, Rabinovici-Cohen S, Hexter E. FuseMedML: a framework for accelerated discovery in machine learning based biomedicine. J Open Source Softw. 2023;8(81):4943. doi: 10.21105/joss.04943.

24. Kirillov A, Mintun E, Ravi N, Mao H, Rolland C, Gustafson L, et al. Segment anything. Proc IEEE/CVF Int Conf Comput Vis. 2023:4015-4026. doi: 10.1109/ICCV51070.2023.00371.

25. Ravi N, Gabeur V, Hu YT, Hu R, Ryali C, Ma T, et al. SAM 2: Segment Anything in Images and Videos. International Conference on Learning Representations. 2025. Available at: https://proceedings.iclr.cc/paper_files/paper/2025/hash/45c1f6a8cbf2da59ebf2c802b4f742cd-Abstract-Conference.html. Accessed September 25, 2026.

26. Yu AS, Hariri M, Zhang X, Yang M, Chaudhary V, Li X. Novel adaptation of video segmentation to 3D MRI: efficient zero-shot knee segmentation with SAM2. arXiv [Preprint]. Published online August 8, 2024. doi: 10.48550/arXiv.2408.04762.

27. Hoyer G, Tong MW, Bhattacharjee R, Pedoia V, Majumdar S. Clinical utility of foundation models in musculoskeletal MRI for biomarker fidelity and predictive outcomes. NPJ Digit Med. 2026;9(1):383. doi: 10.1038/s41746-026-02520-w.

28. Chen T, Kornblith S, Norouzi M, Hinton G. A simple framework for contrastive learning of visual representations. Proc Mach Learn Res. 2020;119:1597-1607. Available at: https://proceedings.mlr.press/v119/chen20j.html. Accessed September 25, 2026.

29. Caron M, Touvron H, Misra I, Jegou H, Mairal J, Bojanowski P, et al. Emerging properties in self-supervised vision transformers. Proc IEEE/CVF Int Conf Comput Vis. 2021:9650-9660. doi: 10.1109/ICCV48922.2021.00951.

30. Dosovitskiy A, Beyer L, Kolesnikov A, Weissenborn D, Zhai X, Unterthiner T, et al. An image is worth 16x16 words: transformers for image recognition at scale. International Conference on Learning Representations. 2021. Available at: https://openreview.net/forum?id=YicbFdNTTy. Accessed September 25, 2026.

31. Chen RJ, Ding T, Lu MY, Williamson DFK, Jaume G, Song AH, et al. Towards a general-purpose foundation model for computational pathology. Nat Med. 2024;30(3):850-862. doi: 10.1038/s41591-024-02857-3.

32. Vorontsov E, Bozkurt A, Casson A, Shaikovski G, Zelechowski M, Severson K, et al. A foundation model for clinical-grade computational pathology and rare cancers detection. Nat Med. 2024;30(10):2924-2935. doi: 10.1038/s41591-024-03141-0.

33. Xu H, Usuyama N, Bagga J, Zhang S, Rao R, Naumann T, et al. A whole-slide foundation model for digital pathology from real-world data. Nature. 2024;630(8015):181-188. doi: 10.1038/s41586-024-07441-w.

34. Tian K, Jiang Y, Diao Q, Lin C, Wang L, Yuan Z. Designing BERT for convolutional networks: sparse and hierarchical masked modeling. International Conference on Learning Representations. 2023.
35. Woo S, Debnath S, Hu R, Chen X, Liu Z, Kweon IS, et al. ConvNeXt V2: co-designing and scaling ConvNets with masked autoencoders. Proc IEEE/CVF Conf Comput Vis Pattern Recognit. 2023:16133-16142. doi: 10.1109/CVPR52729.2023.01548.
36. Zhou Z, Sodha V, Siddiquee MMR, Feng R, Tajbakhsh N, Gotway MB, et al. Models Genesis: generic autodidactic models for 3D medical image analysis. Med Image Comput Comput Assist Interv. 2019;11767:384-393. doi: 10.1007/978-3-030-32251-9_42.
37. Li Y, Wynne JF, Wu Y, Qiu RLJ, Tian S, Wang T, et al. Automatic medical imaging segmentation via self-supervising large-scale convolutional neural networks. Radiother Oncol. 2025;204:110711. doi: 10.1016/j.radonc.2025.110711.
38. Hatamizadeh A, Tang Y, Nath V, Yang D, Myronenko A, Landman B, et al. UNETR: transformers for 3D medical image segmentation. Proc IEEE/CVF Winter Conf Appl Comput Vis. 2022:574-584. doi: 10.1109/WACV51458.2022.00181.
39. Panfilov E, Tiulpin A, Nieminen MT, Saarakkala S, Casula V. Deep learning-based segmentation of knee MRI for fully automatic subregional morphological assessment of cartilage tissues: data from the Osteoarthritis Initiative. J Orthop Res. 2022;40(5):1113-1124. doi: 10.1002/jor.25150.
40. Burton W 2nd, Myers CA, Rullkoetter PJ. Semi-supervised learning for automatic segmentation of the knee from MRI with convolutional neural networks. Comput Methods Programs Biomed. 2020;189:105328. doi: 10.1016/j.cmpb.2020.105328.
41. Zhou Y, Nderitu P, Goh JHL, Engelmann J, Wagner SK, Ran A, et al. Generalist versus specialist vision foundation models for ocular disease and oculomics. arXiv [Preprint]. Published online September 3, 2025. doi: 10.48550/arXiv.2509.03421.
42. Hou Q, Zhou Y, Goh JHL, Zou K, Yew SME, Srinivasan S, et al. Can a natural image-based foundation model outperform a retina-specific model in detecting ocular and systemic diseases? Ophthalmol Sci. 2026;6(1):100923. doi: 10.1016/j.xops.2025.100923.

## Tables

**Table 1. Cohort and imaging characteristics.** Data are reported for the pretraining source and newly added transfer cohorts. Age and sex were not available in the de-identified raw fastMRI data linked to fastMRI+ annotations. OAI pretraining included 19,011 MRI series from 4,791 participants. The fastMRI+ source reports 1,172 knee examinations, each consisting of one MRI series. SKM-TEA includes 155 patients and 155 unique qDESS knee scans; each qDESS acquisition contains two echoes and is counted here as one MRI series. APM PD FS TSE series were sagittal 2D acquisitions whose contiguous slices were stacked for the 3D model input. Abbreviations: APM = Arthroscopic Partial Meniscectomy; CV = cross-validation; DESS = dual-echo steady state; OAI = Osteoarthritis Initiative; PD = proton density; PD FS = proton-density fat-suppressed; qDESS = quantitative dual-echo steady state; SKM-TEA = Stanford Knee MRI with Multi-Task Evaluation; TSE = turbo spin echo.

| Characteristic | OAI | fastMRI+ | SKM-TEA | APM | APM |
|---|---|---|---|---|---|
| Task | Pretraining | Classification | Segmentation | Classification | Segmentation |
| Participants, No. | 4,791 | 1,172 | 155 | 1,716 | 25 |
| MRI series, No. | 19,011 | 1,172 | 155 | 1,716 | 25 |
| Sex | 2,799 female; 1,992 male | Not available | 65 female; 90 male | 848 female; 868 male | 12 female; 13 male |
| Median age, y | 65 | Not available | 42 | 57 | 55 |
| Age range, y | 48-88 | Not available | 16-87 | 45-90 | 46-72 |
| Scanner models | Siemens Trio | Siemens systems | Two GE Discovery MR750 systems | Siemens and Philips (15 models) | Siemens (6 models) |
| Field strength, T | 3 | 1.5 and 3 | 3 | 1, 1.5, and 3 | 1.5 and 3 |
| Sequences | DESS (n=9,493); TSE (n=9,518) | PD and PD FS | qDESS | PD FS (n=1,429); dual PD/T2 FS (n=287) | PD FS (n=17); dual PD/T2 FS (n=8) |
| Analysis split/design | 19,011 series; 5% self-supervised validation | 703 training; 234 validation; 235 test | 86 training; 33 validation; 36 test | 1,028 training; 345 validation; 343 test | Five-fold CV with 5 fixed held-out test cases |

**Table 2. fastMRI+ two-label classification performance on the held-out test set.** Values are macro estimates across meniscus tear and cartilage defect, with patient-level bootstrap 95% confidence intervals. Abbreviations: CI = confidence interval; PR AUC = precision-recall area under the curve; ROC AUC = receiver operating characteristic area under the curve.

| Training cases (%) | Initial weights | Macro balanced accuracy (95% CI) | Macro F1 (95% CI) | Macro ROC AUC (95% CI) | Macro PR AUC (95% CI) |
|---|---|---|---|---|---|
| 703 (100%) | Random | 0.666 (0.618-0.710) | 0.706 (0.654-0.751) | 0.714 (0.663-0.763) | 0.723 (0.651-0.791) |
| 703 (100%) | SuPreM | 0.637 (0.589-0.683) | 0.624 (0.563-0.680) | 0.686 (0.633-0.737) | 0.701 (0.631-0.769) |
| 703 (100%) | KneePreM | 0.701 (0.653-0.745) | 0.716 (0.659-0.766) | 0.777 (0.727-0.822) | 0.788 (0.717-0.846) |
| 492 (70%) | Random | 0.641 (0.599-0.689) | 0.660 (0.606-0.708) | 0.695 (0.643-0.749) | 0.703 (0.632-0.778) |
| 492 (70%) | SuPreM | 0.613 (0.568-0.661) | 0.670 (0.613-0.718) | 0.669 (0.615-0.726) | 0.693 (0.622-0.764) |
| 492 (70%) | KneePreM | 0.701 (0.661-0.739) | 0.700 (0.652-0.749) | 0.748 (0.698-0.789) | 0.774 (0.712-0.827) |
| 352 (50%) | Random | 0.632 (0.584-0.680) | 0.649 (0.590-0.703) | 0.672 (0.614-0.724) | 0.699 (0.623-0.769) |
| 352 (50%) | SuPreM | 0.593 (0.546-0.639) | 0.620 (0.559-0.672) | 0.629 (0.573-0.686) | 0.669 (0.592-0.748) |
| 352 (50%) | KneePreM | 0.660 (0.616-0.706) | 0.669 (0.613-0.727) | 0.722 (0.675-0.771) | 0.770 (0.709-0.826) |
| 211 (30%) | Random | 0.601 (0.554-0.649) | 0.595 (0.529-0.655) | 0.636 (0.582-0.692) | 0.668 (0.601-0.742) |
| 211 (30%) | SuPreM | 0.534 (0.498-0.572) | 0.506 (0.454-0.560) | 0.579 (0.524-0.636) | 0.595 (0.528-0.665) |
| 211 (30%) | KneePreM | 0.648 (0.603-0.689) | 0.659 (0.599-0.714) | 0.702 (0.653-0.750) | 0.738 (0.675-0.796) |

**Table 3. APM eight-target compartment-status classification performance on the held-out test set.** Values are macro estimates across eight compartment-status targets, with patient-level bootstrap 95% confidence intervals. Abbreviations: CI = confidence interval; PR AUC = precision-recall area under the curve; ROC AUC = receiver operating characteristic area under the curve.

| Training cases (%) | Initial weights | Macro balanced accuracy (95% CI) | Macro F1 (95% CI) | Macro ROC AUC (95% CI) | Macro PR AUC (95% CI) |
|---|---|---|---|---|---|
| 1028 (100%) | Random | 0.658 (0.635-0.679) | 0.687 (0.660-0.713) | 0.741 (0.717-0.765) | 0.739 (0.707-0.770) |
| 1028 (100%) | SuPreM | 0.610 (0.588-0.632) | 0.618 (0.586-0.647) | 0.688 (0.660-0.714) | 0.691 (0.658-0.725) |
| 1028 (100%) | KneePreM | 0.686 (0.664-0.708) | 0.704 (0.678-0.730) | 0.787 (0.763-0.808) | 0.781 (0.750-0.812) |
| 720 (70%) | Random | 0.636 (0.612-0.660) | 0.662 (0.633-0.691) | 0.708 (0.680-0.734) | 0.693 (0.658-0.726) |
| 720 (70%) | SuPreM | 0.571 (0.550-0.592) | 0.616 (0.588-0.643) | 0.633 (0.604-0.661) | 0.636 (0.600-0.670) |
| 720 (70%) | KneePreM | 0.636 (0.616-0.658) | 0.689 (0.664-0.715) | 0.740 (0.716-0.763) | 0.733 (0.705-0.764) |
| 514 (50%) | Random | 0.597 (0.575-0.620) | 0.548 (0.513-0.581) | 0.659 (0.630-0.689) | 0.661 (0.626-0.696) |
| 514 (50%) | SuPreM | 0.535 (0.521-0.552) | 0.553 (0.528-0.581) | 0.593 (0.562-0.622) | 0.592 (0.558-0.625) |
| 514 (50%) | KneePreM | 0.631 (0.608-0.655) | 0.674 (0.647-0.702) | 0.718 (0.692-0.745) | 0.726 (0.695-0.756) |
| 308 (30%) | Random | 0.547 (0.528-0.566) | 0.570 (0.545-0.595) | 0.614 (0.584-0.644) | 0.614 (0.581-0.647) |
| 308 (30%) | SuPreM | 0.527 (0.507-0.547) | 0.516 (0.486-0.549) | 0.586 (0.554-0.614) | 0.592 (0.558-0.625) |
| 308 (30%) | KneePreM | 0.606 (0.583-0.628) | 0.609 (0.580-0.636) | 0.663 (0.636-0.690) | 0.649 (0.618-0.680) |

**Table 4. SKM-TEA overall and structure-level Dice by training-set size.** Abbreviations: FC = femoral cartilage; LM = lateral meniscus; LTC = lateral tibial cartilage; MM = medial meniscus; MTC = medial tibial cartilage; PC = patellar cartilage.

| Training cases (%) | Initial weights | Mean Dice | FC | LTC | MTC | PC | LM | MM |
|---|---|---|---|---|---|---|---|---|
| 86 (100%) | Random | 0.835 | 0.852 | 0.842 | 0.838 | 0.833 | 0.819 | 0.825 |
| 86 (100%) | SuPreM | 0.840 | 0.855 | 0.845 | 0.839 | 0.850 | 0.830 | 0.824 |
| 86 (100%) | KneePreM | 0.842 | 0.850 | 0.848 | 0.843 | 0.845 | 0.836 | 0.833 |
| 60 (70%) | Random | 0.827 | 0.849 | 0.843 | 0.835 | 0.824 | 0.801 | 0.808 |
| 60 (70%) | SuPreM | 0.832 | 0.852 | 0.844 | 0.838 | 0.837 | 0.814 | 0.810 |
| 60 (70%) | KneePreM | 0.838 | 0.855 | 0.845 | 0.837 | 0.845 | 0.825 | 0.822 |
| 26 (30%) | Random | 0.802 | 0.823 | 0.828 | 0.813 | 0.798 | 0.774 | 0.773 |
| 26 (30%) | SuPreM | 0.812 | 0.841 | 0.821 | 0.806 | 0.827 | 0.793 | 0.786 |
| 26 (30%) | KneePreM | 0.826 | 0.836 | 0.840 | 0.829 | 0.826 | 0.814 | 0.808 |
| 13 (15%) | Random | 0.737 | 0.815 | 0.729 | 0.694 | 0.755 | 0.721 | 0.705 |
| 13 (15%) | SuPreM | 0.752 | 0.827 | 0.743 | 0.748 | 0.785 | 0.705 | 0.707 |
| 13 (15%) | KneePreM | 0.779 | 0.818 | 0.785 | 0.795 | 0.779 | 0.752 | 0.746 |
| 4 (5%) | Random | 0.555 | 0.745 | 0.491 | 0.472 | 0.643 | 0.539 | 0.440 |
| 4 (5%) | SuPreM | 0.617 | 0.786 | 0.587 | 0.556 | 0.691 | 0.597 | 0.484 |
| 4 (5%) | KneePreM | 0.628 | 0.767 | 0.620 | 0.634 | 0.700 | 0.552 | 0.494 |

**Table 5. APM clinical segmentation overall and structure-level Dice by training-set size.** Abbreviations: FC = femoral cartilage; LM = lateral meniscus; LTC = lateral tibial cartilage; MM = medial meniscus; MTC = medial tibial cartilage; PC = patellar cartilage

| Training cases (%) | Initial weights | Mean Dice | FC | LTC | MTC | PC | LM | MM |
|---|---|---|---|---|---|---|---|---|
| 16 (100%) | Random | 0.731 | 0.741 | 0.701 | 0.680 | 0.774 | 0.775 | 0.715 |
| 16 (100%) | SuPreM | 0.752 | 0.753 | 0.730 | 0.712 | 0.788 | 0.778 | 0.751 |
| 16 (100%) | KneePreM | 0.779 | 0.762 | 0.765 | 0.737 | 0.810 | 0.822 | 0.778 |
| 12(75%) | Random | 0.672 | 0.724 | 0.614 | 0.633 | 0.730 | 0.688 | 0.643 |
| 12 (75%) | SuPreM | 0.713 | 0.752 | 0.665 | 0.658 | 0.792 | 0.729 | 0.683 |
| 12(75%) | KneePreM | 0.746 | 0.743 | 0.737 | 0.714 | 0.762 | 0.782 | 0.737 |
| 8(50%) | Random | 0.617 | 0.704 | 0.534 | 0.575 | 0.714 | 0.610 | 0.567 |
| 8(50%) | SuPreM | 0.664 | 0.727 | 0.582 | 0.636 | 0.769 | 0.632 | 0.636 |
| 8(50%) | KneePreM | 0.684 | 0.727 | 0.628 | 0.660 | 0.719 | 0.698 | 0.675 |
| 4(25%) | Random | 0.503 | 0.630 | 0.418 | 0.431 | 0.624 | 0.501 | 0.413 |
| 4(25%) | SuPreM | 0.533 | 0.673 | 0.410 | 0.418 | 0.709 | 0.526 | 0.464 |
| 4(25%) | KneePreM | 0.578 | 0.675 | 0.496 | 0.486 | 0.670 | 0.602 | 0.540 |

## Figure Legends

**Figure 1.** Overview of the KneePreM framework. Illustrative 3D MRI volumes indicate the inputs and expected outputs at each stage. KneePreM is based on a standard 3D U-Net architecture with four resolution levels. During pretraining (a), the full encoder and decoder are trained on Osteoarthritis Initiative data using a reconstruction objective with masked MRI volumes as input. The encoder weights learned during pretraining are transferred to downstream tasks and fine-tuned. During downstream classification (b), the pretrained encoder is coupled with one or more fully connected classification heads, each producing a presence-or-absence prediction for a specific target. During downstream segmentation (c), the pretrained encoder is paired with a randomly initialized 3D U-Net decoder that serves as the segmentation head.

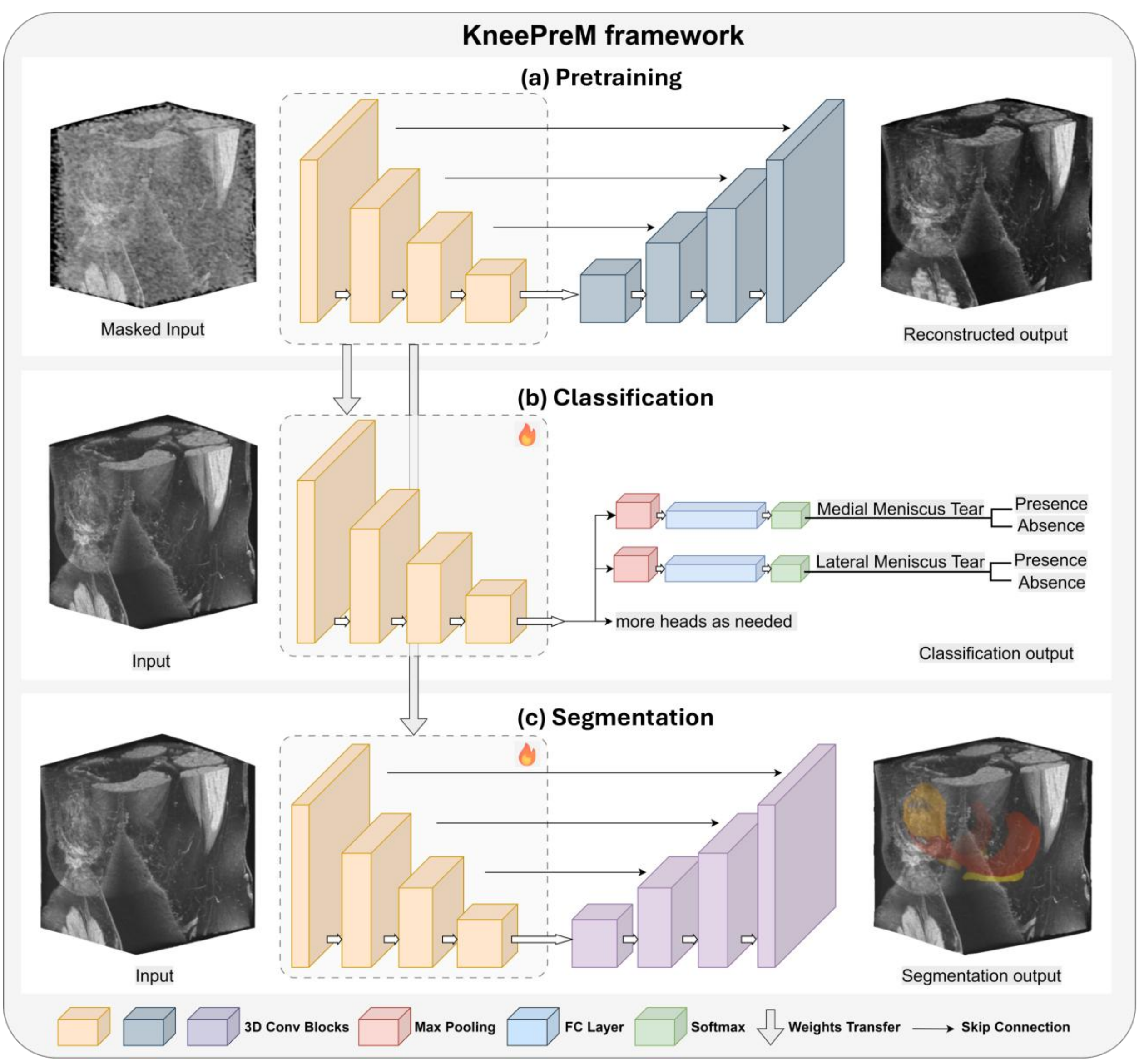

**Figure 2.** Classification performance for fastMRI+ and APM. The top row shows fastMRI+ classification: ROC AUC by training-set size (a1), full-data bootstrap ROC AUC distributions with pairwise p values (a2), and full-data ROC curves (a3). The bottom row shows the corresponding macro ROC AUC analyses for the eight-target APM classification task: training-set size (b1), full-data bootstrap distributions (b2), and full-data ROC curves (b3). Scratch denotes random initialization.

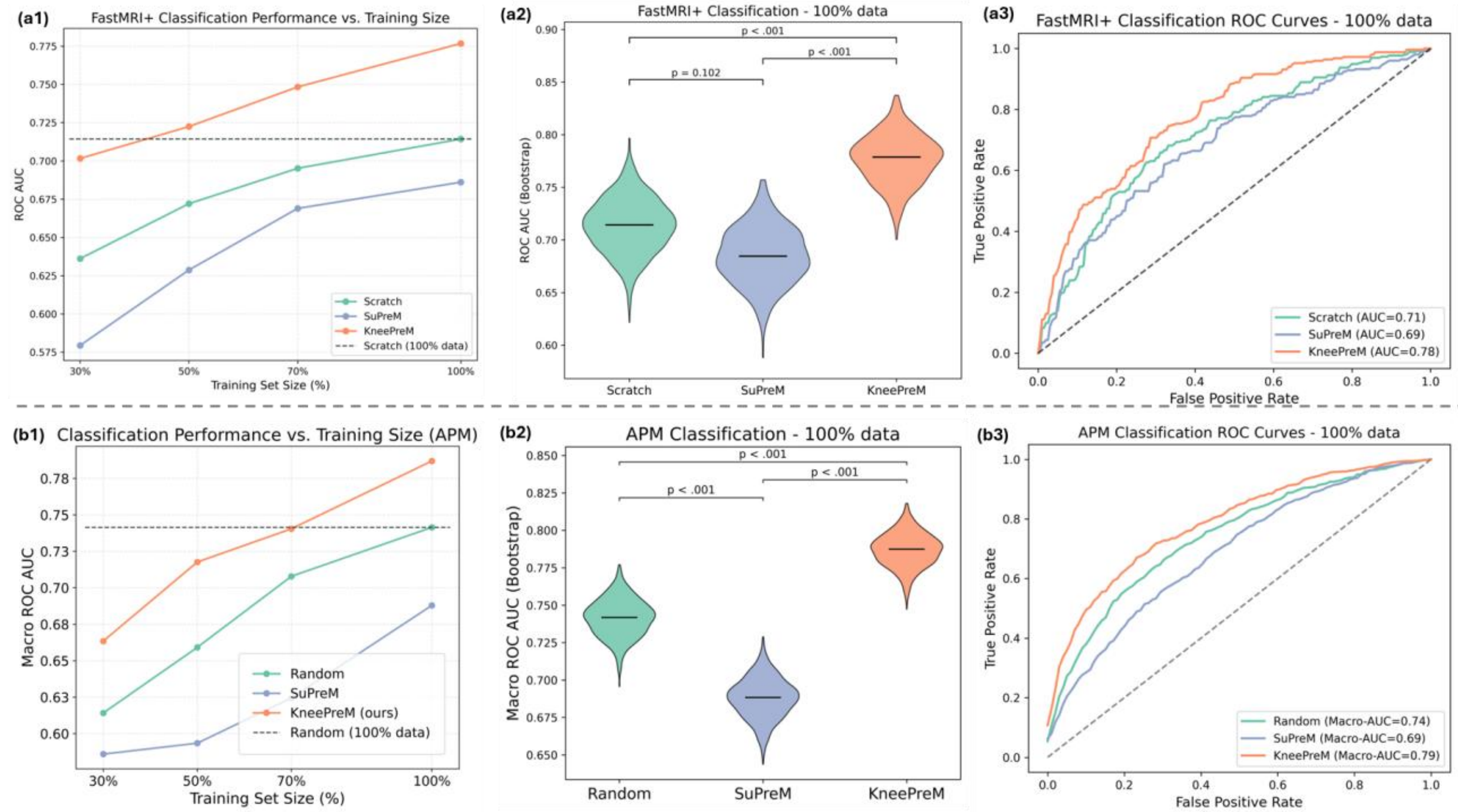

**Figure 3.** Full-data classification confusion matrices. The upper block (a) contains the meniscus tear matrices at left and cartilage defect matrices at right with random, SuPreM, and KneePreM initialization. The lower block (b) contains APM confusion matrices for lateral and medial meniscus status, lateral and medial femoral condyle status, lateral and medial tibial plateau status, patellar cartilage status, and trochlear status with random, SuPreM, and KneePreM initialization. Rows indicate reference classes and columns indicate predicted classes. Scratch denotes random initialization.

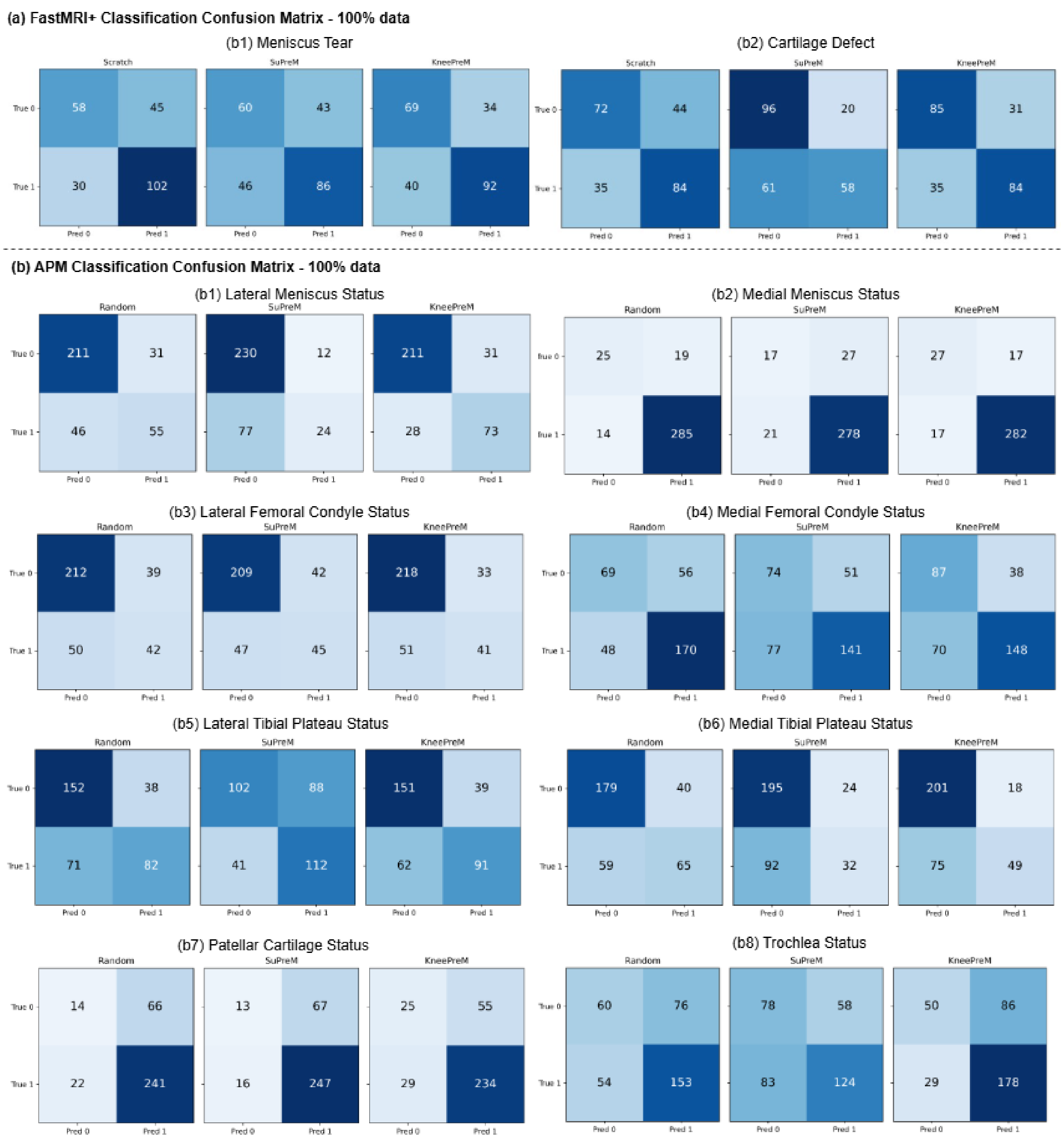

**Figure 4.** Segmentation performance for SKM-TEA and APM. The top row shows SKM-TEA mean Dice by training-set size (a1) and examination-level Dice distributions at 70% (a2) and 100% (a3) training data. The bottom row shows APM mean Dice by training-set size (b1) and paired fold-level Dice distributions at 75% (b2) and 100% (b3) training data. Pairwise p values and KneePreM-minus-random effect summaries are displayed in the distribution panels.

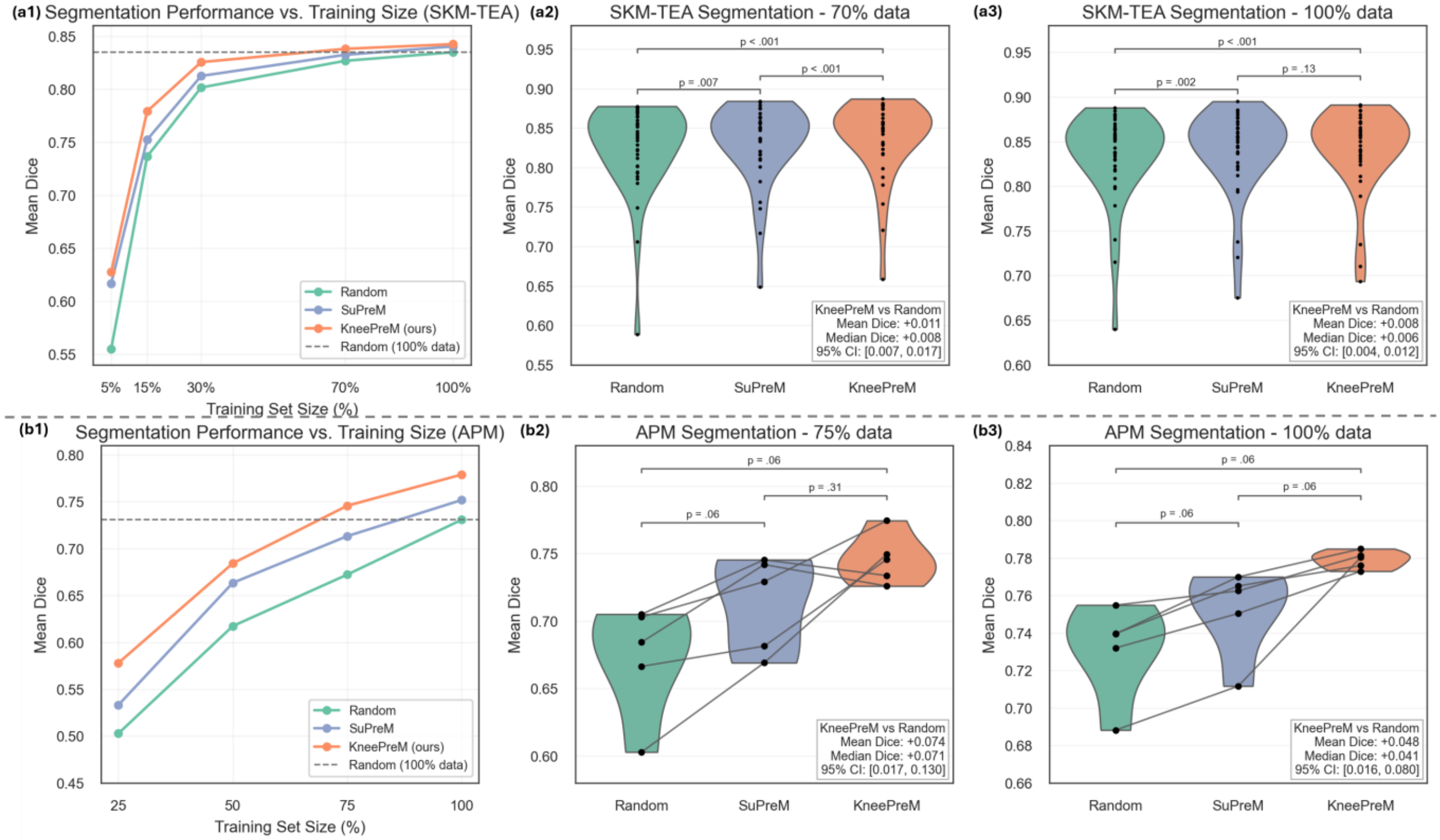

**Figure 5.** Qualitative comparison of knee MRI segmentation in SKM-TEA (a–d) and APM (e–h). Each row shows one sagittal slice; columns show the input image, reference standard, and predictions from Random initialization, SuPreM, and KneePreM, respectively. SKM-TEA examples use 100% (a, b) or 70% (c, d) of the training data; APM examples use 100% (e, f) or 75% (g–h). Overlays indicate FC (red), LTC (green), MTC (blue), PC (yellow), LM (magenta), and MM (cyan).

Yellow dashed circles identify selected baseline errors. In SKM-TEA, Random and SuPreM incompletely delineate LTC in (a), LM in (b, c), and MM in (d). In APM, the baseline methods incompletely delineate MM in (e). In (f), Random mislabels part of the FC region as PC, while SuPreM incompletely delineates MM. The highlighted errors involve LTC in (g) and FC in (h). KneePreM more closely matches the reference in these regions. Values shown on prediction panels are compartment-specific slice-wise Dice scores.

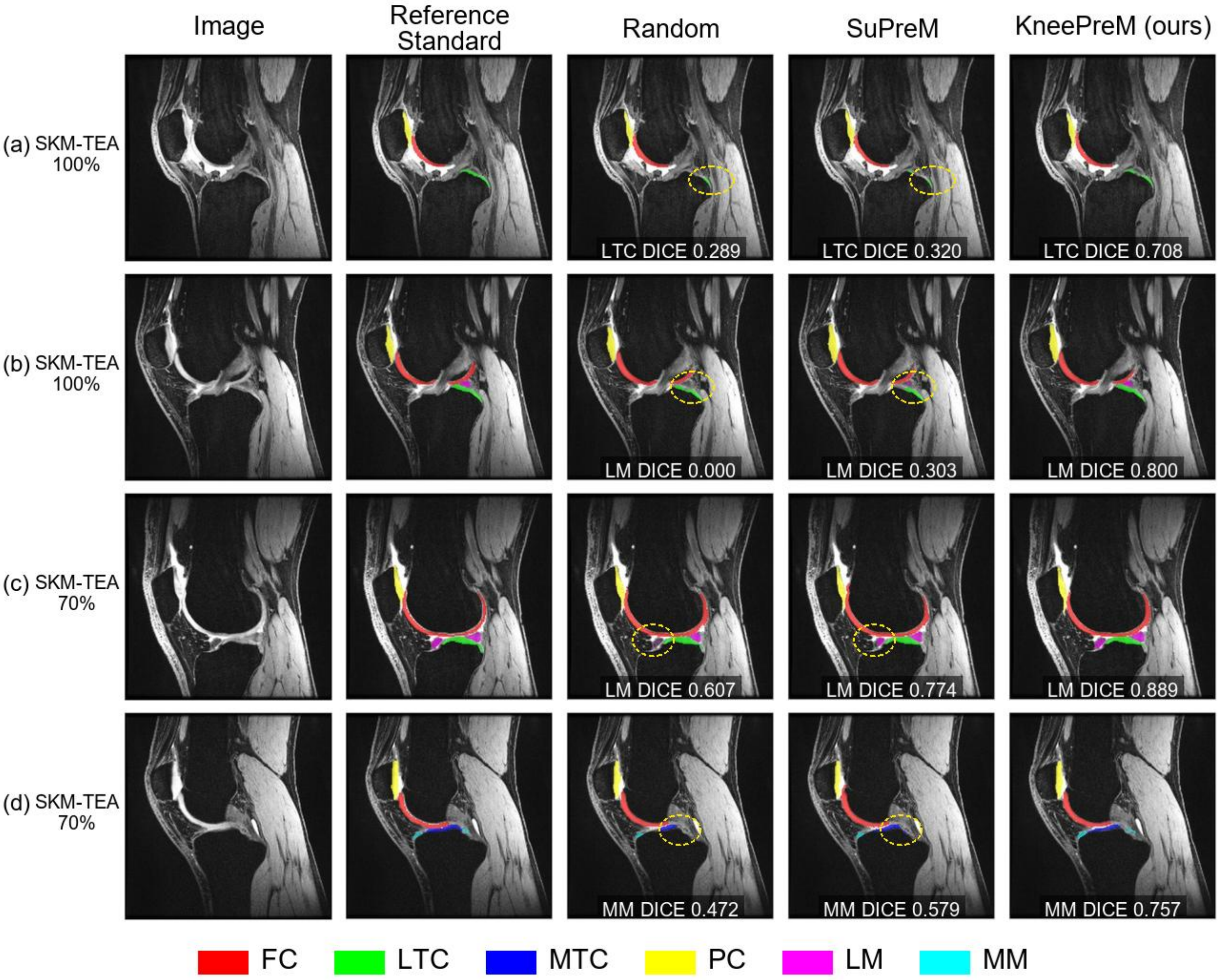

Image
Reference Standard
Random
SuPreM
KneePreM (ours)
(a) SKM-TEA 100%
(b) SKM-TEA 100%
(c) SKM-TEA 70%
(d) SKM-TEA 70%
LTC DICE 0.289
LTC DICE 0.320
LTC DICE 0.708
LM DICE 0.000
LM DICE 0.303
LM DICE 0.800
LM DICE 0.607
LM DICE 0.774
LM DICE 0.889
MM DICE 0.472
MM DICE 0.579
MM DICE 0.757
FC
LTC
MTC
PC
LM
MM

**Figure 5 (continued).**

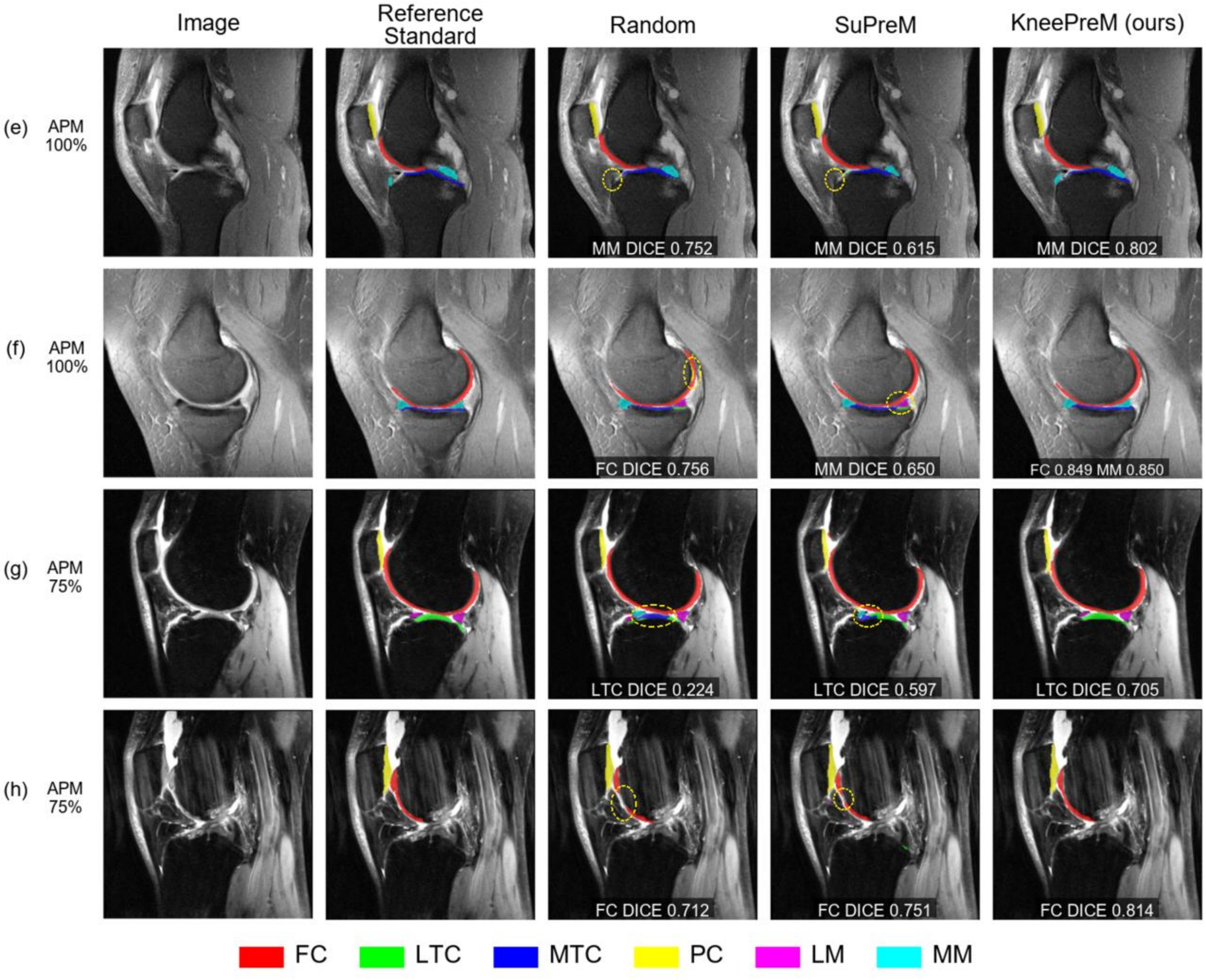